\documentclass[10pt,journal,compsoc]{IEEEtran}

\usepackage[]{algorithm2e}
\usepackage{amsmath}
\usepackage{amssymb}
\usepackage{mathtools,xparse}
\PassOptionsToPackage{hyphens}{url}
\usepackage{hyperref}
\usepackage{url}

\usepackage{authblk}

\DeclareMathOperator*{\argmin}{arg\,min}
\DeclareMathOperator*{\sign}{sign}
\DeclarePairedDelimiter{\abs}{\lvert}{\rvert}

\begin{document}

\title{Deep Learning Stereo Vision at the edge}

\author[1]{Luca Puglia}
\author[1]{Cormac Brick}
\affil[1]{Intel R\&D, \textit{name.surname@intel.com}}

\IEEEtitleabstractindextext{%
\begin{abstract}

We present an overview of the methodology used to build a new stereo vision solution that is suitable for System on Chip. This new solution was developed to bring computer vision capability to embedded devices that live in a power constrained environment. The solution is constructured as a hybrid between classical Stereo Vision techniques and deep learning approaches. The stereoscopic module is composed of two separate modules: one that accelerates the neural network we trained and one that accelerates the front-end part. The system is completely passive and does not require any structured light to obtain very compelling accuracy. With respect to the previous Stereo Vision solutions offered by the industries we offer a major improvement is robustness to noise. This is mainly possible due to the deep learning part of the chosen architecture. We submitted our result to Middlebury dataset challenge. It currently ranks as the best System on Chip solution. The system has been developed for low latency applications which require better than real time performance on high definition videos.

\end{abstract}

}

\maketitle

\IEEEdisplaynontitleabstractindextext

\IEEEpeerreviewmaketitle

\ifCLASSOPTIONcompsoc
\IEEEraisesectionheading{\section{Introduction}\label{sec:introduction}}
\else
\section{Introduction}
\label{sec:introduction}
\fi
In the past few years the demand for accurate, low latency, high definition 3D environment reconstruction has increased dramatically. Many existing applications can benefit from the addition of scene depth information. Furthermore, the existence of high quality 3D reconstruction liberates some completely new use cases (e.g. Augmented Reality with occlusion handling). The output of a 3D reconstruction pipeline is usually either a 3D point cloud or a depth map (a.k.a. disparity map). These representations can be used alongside normal images to make many applications far more robust. A few use cases that can benefit from the addition of depth information are: face recognition, object count, object segmentation and autonomous driving. The fundamental requirements of these systems is to provide high accuracy at a real time frame rate ($>$30 fps), high resolution ($>$720p) and very low latency. The only way to fulfill all these requirements at the same time is to do the necessary computation on board. The only current method to reach this kind of performance using off-the-shelf solutions is to use GPU based systems. This kind of hardware, though, is not ideal for a power constrained environment (i.e.  surveillance cameras in the wild). For this reason we developed a custom System on Chip (SoC) capable of achieving industry standard state of the art.

In general, there are two different classes of 3D reconstruction techniques. On class uses an "active" approach that couple an optic sensor with an optic emitter: while the emitter projects a known pattern on the scene, the sensor infer the environment from the deformation of the pattern \cite{zhang2012microsoft}. The other class uses a "passive" approach using only optic sensors for 3D reconstruction. This can be challenging and more sophisticated techniques (which requires more computation) are required. However in return, the main advantages are a lower power usage (mainly because no emitter is used) and no interference problems (mainly given by the sun light). The most established and reliable algorithms for passive reconstruction are without any doubt Stereo Vision algorithms. These techniques try to solve the \textit{stereo matching problem}: given a pair of frames taken from two different twin cameras, match every pixel from the first view with a pixel from the second. Some constraints can be added to the camera positions to simplify the problem. In particular the focal axes of the two cameras are considered. In fact, if the two axes are parallel, the search for the correct pixel match is much easier. In this set up the Stereo Vision algorithm has to search for the correct match on a single line of pixels instead of the whole image. Unfortunately the perfect alignment of camera axis can only be accurate to some extent. For this reason a further process of camera calibration and image rectification is used to make sure that the epipolar lines in the two input frames are parallel. Once that the input frames are captured and rectified they are processed through a series of logical steps retrieving the 3D geometry of the scene. The output of this process is a third frame whose pixels contain information about the depth of the scene.

A taxonomy of stereo vision approaches has been described in \cite{SchaSze}. Generally speaking a classical pipeline is composed of four main stages:

\begin{itemize}
    \item Cost Matching Computation;
    \item Cost Aggregation;
    \item Disparity Selection;
    \item Disparity Refinement.
\end{itemize}

\begin{figure*}
    \centering
    \includegraphics[trim={0 20px 0 18px},clip]{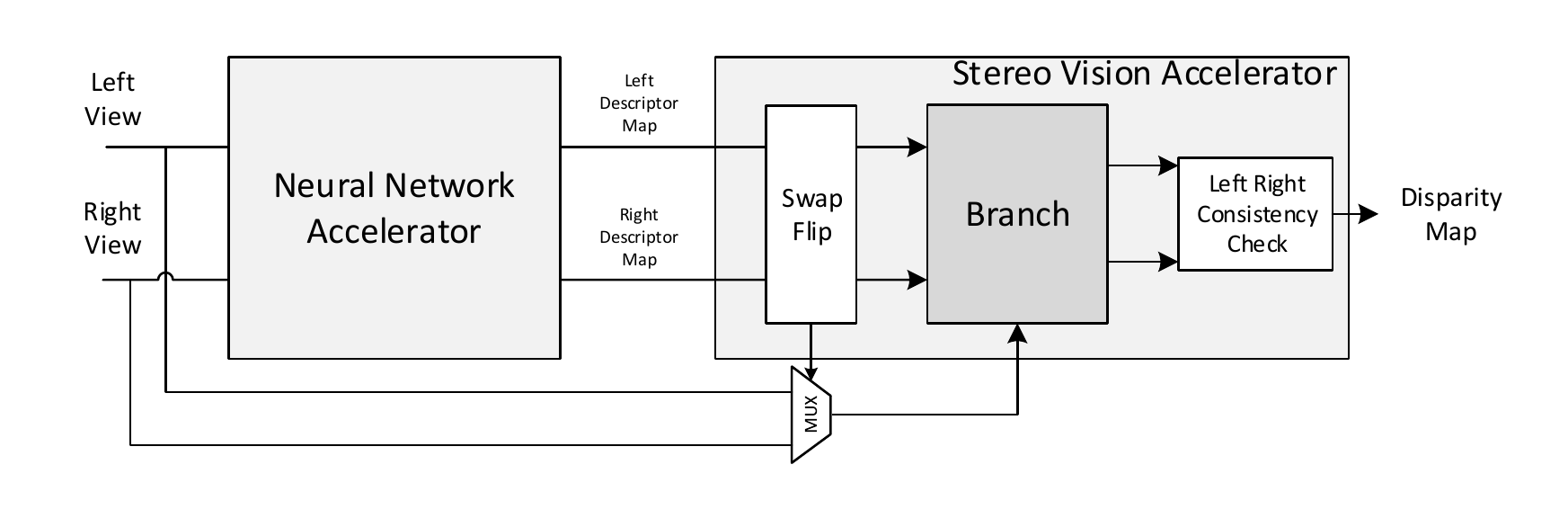}
    \caption{The architecture is composed by two separate modules, the Neural Network Acclerator and the Stereo Vision Accelerator, the first is in charge of computing the descriptor of the left and right views, these are then used by the main module to compute the final disparity map.}
    \label{fig:arch}
\end{figure*}

In order to run the first stage (the Cost Matching Computation) we need to define the cost of the matching Essentially we define a distance function able to compare one pixel from one frame against one from the other frame. The smaller the distance the lower the matching cost. A low matching cost means that two pixel are very similar, this similarity can be defined in many way. The most naive way is to compare the color of the two pixels. Unfortunately two pixels can have similar color but not belong to the same spot in the scene. For this reason more robust similarity functions have been proposed. Most of these are based on sliding window approaches, where a signature is extracted from every sub-window in the image. A function that is able to map one feature vector (every value in the window) to a different feature space is called an embedding function. The most famous embedding function in Stereo Vision field is the census transform \cite{Zabih}. In this hand crafted technique all the pixel values in a fixed size window are compared with the value of the pixel at the center. If a pixel has a greater value than the center we place a 1 in the final binary vector, 0 otherwise, in this way a signature (also called descriptor) for every sub-window is created.

Once a cost function is defined, the Cost Matching Computation step is used to combine the two frames into a volume: each descriptor of position $(x,y)$ in the reference frame is compared with $N$ descriptors from the target frame. The positions of the target pixels belong to the range [$(x-N,y)$, $(x,y)$]. The output volume has the same height and width of the input frames and depth $N$. The depth size has to be chosen carefully: a big value for $N$ increases drastically the memory requirements of the volume. This volume contains the information of $N$ comparisons for each $(x,y)$ position in the reference frame. In an ideal world the $argmin$ of each $N$ vector will represent the correct displacement (or disparity) for each final disparity map pixel. Unfortunately this step can be very noisy because many points can have similar a descriptor but not represent the same object, for this reason the Cost Aggregation step is fundamental to make the volume information more reliable. The values in the volume are modified using a set of specific rules and their neighboring values. Finally in the Disparity Selection step the $argmin$ for each position $(x,y)$ in the volume is used to chose the best disparity. The Disparity Refinement step is used to increase the overall accuracy using smoothing and denoising filters. Even though standard Stereo Vision is generally performed in a passive way, it is possible to project a pattern on the scene to make the matching process more robust \cite{Keselman2017IntelRS}.

With the beginning of the Deep Learning era many of these steps (if not all) have been replaced by deep neural network \cite{Luo, Kendall, Fanello}. This new tool in the hand of researchers has greatly increased the accuracy of the state-of-the-art approaches. The main drawback of these techniques, though, is the high demand of computational power (billions, if not trillions of operations) and memory footprints (Gigabytes of storage for each pair of frames). Such requirements are too high for efficient SoC solutions. For this reason we had to find the right trade-off between accuracy and performance.
Our stereo vision module is able to deliver high accuracy disparity maps with very low latency and low power usage, the rest of the paper is divided in the following sections:

\begin{itemize}
\item Section 2 expand the architecture of the Stereo Vision module;
\item Section 3 lists the of problems to address in order to accelerate a neural network model;
\item Section 4 describes the motivation for the usage of binary descriptors;
\item Section 5 shows the ablation study we used to determined the best model to use;
\item Section 6 expands the quantization technique we used to compress the network;
\item Section 7 and 8 shows the results and conclusion drawn.
\end{itemize}

\section{The architecture}

The main difference with off-the-shelf solution is the usage of a deep learning based descriptor. It enabled the removal of the census transform module which has very well known limitations. Figure:~\ref{fig:arch} shows an accurate representation of the new architecture: on the left side the Neural Network Acclerator (NNA) module is used to accelerate the computation of the neural network descriptor. The output of the NNA is then fed to the Stereo Vision Accelerator for the computation of the final disparity map.
A technique that increases the robustness of a Stereo Vision algorithm is the \textit{left right consistency check}: the basic Stereo Vision algorithm uses the left image as a reference frame and the right one as target. This means that the pixel from left is compared with multiple pixels on the right. There is no reason to choose one of the two frames over the other as reference or target. In fact, most of the classical approaches compute both disparity maps, left vs right and right vs left. The two disparity maps are then checked for consistency and merged together. This improves the final accuracy a lot and very easily removes random fluctuations in the disparity values. However it effectively reduces in half the frame rate of the pipeline. We addressed this problem in the stereo vision accelerator and decide to make the NNA run a single time for each stereo pair. For the core of the accelerator we decided to go for a very basic box filter that smooths the volume values followed by the very well known Semi Global Matching (SGM) algorithm \cite{Hirschmuller}.

The time complexity of these algorithms is generally quadratic with the size of the images. For this reason they are not very efficient in single core systems. To increase the throughput of our solution we created a fully pipelined architecture. Once all the sub-modules fill the internal buffers (initial delay) we get an output disparity every two clock ticks (constant complexity).

\section{Deep learning for Stereo Vision}

Deep Learning can be used to greatly enhance the performance of a Stereo Vision pipelines. Many papers in the past few years have shown state-of-the-art results for this particular field. The first proposed technique \cite{zbontar15} used a Convolutional Neural Network (CNN) called MC-CNN as a first stage of a classical Stereo Vision pipeline. The convolutional kernels are used in a sliding window fashion in order to obtain a 32 or 64 floating point descriptor for every sub-window. The descriptors are then compared in different way according to the modality chosen. Two versions are proposed:
\begin{itemize}
    \item An accurate one in which another CNN is used to learn a Cost Matching function;
    \item A fast one that uses a simple dot product (a.k.a. cosine similarity) as the Cost Matching function between the descriptors.
\end{itemize}
Although the computational requirements are very different, the accuracy drop between the two is just around 1\% on common Stereo Vision datasets. When it was first published this method has proven to be a breakthrough in the Stereo Vision community. The ability of CNN to learn new robust descriptors has increased the accuracy a lot over the previous algorithms using handcrafted descriptors. At the time of first submission both KITTI \cite{Geiger2012CVPR} and Middlebury \cite{Scharstein14} tests have ranked it among the top methods. The authors reported a run-time of less than a second for the fast architecture on a GPU.

After this outstanding result many more deep learning based approaches have been proposed. Some of these replace even more steps of the pipeline \cite{Luo}, while others have gone as far as learning the whole 3D reconstruction function end-to-end. We can find example of both passive \cite{Kendall} and active \cite{Fanello} methodology.

\subsection{At the edge acceleration}

In the past few years a lot of new at-the-edge hardware accelerator have been presented both by the academic world \cite{DianNao,eyeriss} and industries \cite{myriadx,tpu,snap}. For our Stereo Vision solution we use the neural network accelerator already present in the chip, namely the NNA. A critical limitation for all embedded devices is the resources available: principally the computational capabilities and memory size. Strictly speaking, the main limitation for CNN accelerators is the quantity of fast memory available to the processor. While a DDR like memory would be big enough to store whole neural network models and their respective feature maps, a cache-like fast-access memory is usually very small in size, in the best cases around the Megabytes order. This may represent a problem for a pipelined architecture that needs to be full the whole time to avoid reduction in efficiency.

Let's take a closer look at the fast architecture proposed in \cite{zbontar15} (by far one of the leanest). The main CNN is a very shallow network composed of only four convolutional layers Table:~\ref{tab:4l}. Since every layer has a 3x3 kernel the final receptive field is 9x9. The final output of the convolution is a feature map that has more or less the same resolution of the input (not considering padding) and 32 channels for each element. Essentially 32 floating point numbers are extracted from each 9x9 patch of the input frame. For this reason there is no constraint on the input resolution and can be changed at will during inference time.
With these basic considerations we can see that the memory and computation required by this simple network is proportional to the size of the input resolution.

\begin{table}[]
\begin{tabular}{l|r|r|r|r|}
\cline{2-5}
                                & \multicolumn{1}{l|}{In. Ch.} & \multicolumn{1}{l|}{Out. Ch.} & \multicolumn{1}{l|}{Kernel} & \multicolumn{1}{l|}{Receptive Field} \\ \hline
\multicolumn{1}{|l|}{ConvReLU1} & 1                            & 32                            & 3                           & 3x3                                  \\ \hline
\multicolumn{1}{|l|}{ConvReLU2} & 32                           & 32                            & 3                           & 5x5                                  \\ \hline
\multicolumn{1}{|l|}{ConvReLU3} & 32                           & 32                            & 3                           & 7x7                                  \\ \hline
\multicolumn{1}{|l|}{Conv4} & 32                           & 32                            & 3                           & 9x9                                  \\ \hline
\end{tabular}

\caption{Structure of the main neural network in \cite{zbontar15} (MC-CNN), the network is very shallow and it consists of only Convolution and ReLU layers.}
\label{tab:4l}
\end{table}

Usually the neural network accelerator uses its fast-access memory to store both the weights of the model and the intermediate feature maps. Even though the model is very shallow the amount of weights to store is:

\begin{equation}
\begin{gathered}
\#weights = 1 * 32 * 3 * 3 + \\
32 * 32 * 3 * 3  +\\
32 * 32 * 3 * 3  +\\
32 * 32 * 3 * 3  = 27,936
\end{gathered}
\end{equation}

If we do not consider any compression technique we may need to store up to 112KB of floating point numbers (float32). Although these figures seem reasonable the main problem with the four layers model is the memory required to store the intermediate feature maps. Even if we assume that we keep in memory the output of each layer at a time we have to store a feature map that can be as big as 1280 * 720 * 32 = 29.5 * 4Byte = 118 MB (again using float32 numbers). This amount is definitively too much for a fast-access (cache-like) memory that is accessed by the neural network accelerator.

Besides the memory problem we have also to consider the actual amount of operations required to process a single frame. The metric that is generally used to profile the amount of operation required by a CNN is the number of FLoating point OPerations (FLOPs), practically speaking the number of multiplication to perform on the input and intermediate feature maps in order to obtain the output, for a single layer we have:

$$\#FLOPs = ch_{in} * ch_{out} * k^2 * input_w * input_h$$

were $ch_{in}$ and $ch_{out}$ are respectively the number of input and output channel, $k$ is the size of the kernel and $input_w$ and $input_h$ are the width and height of the input feature map.
Let's consider the CNN used in \cite{zbontar15} and a 720p HD input frame (i.e. $1280\times720$): we have four layers with input and output channels equals to 32 (except for the first input channel which is 1), all together:

\begin{equation}
\begin{gathered}
\#FLOPs = 1 * 32 * 9 * 1280 * 720 + \\
32 * 32 * 9 * 1280 * 720 +\\
32 * 32 * 9 * 1280 * 720 +\\
32 * 32 * 9 * 1280 * 720 = 34\times 10^9
\end{gathered}
\end{equation}

This result has to be further multiplied by the number of frames per second we want to obtain. No embedded hardware accelerator in the market can reach this kind of performance. For this reason we searched for an even smaller architecture.

The second problem to address is the complexity of the comparison function used for the descriptor. In the original fast architecture the cosine similarity is used. This function is not really hardware friendly (a lot of multiplications and divisions have to be performed). For this reason we decided in favor of a binary descriptor. These kind of descriptor are much more compact and can be compared with a distance function as efficient as the hamming distance.

Finally the third problem to address is the computational demand of floating point multiplications. All the current at-the-edge hardware accelerators do not encourage the usage of floating point multiplication. In fact, it has been shown that 8bit quantization can be used to fully replace full precision model with very low loss \cite{pytorchq, tensorrt}. This not only decreases the footprint of the model in the memory but it also allows the usage of smaller and much more efficient 8 bit multiplier.

\section{Binary descriptors}

\begin{figure}
    \includegraphics[width=\columnwidth]{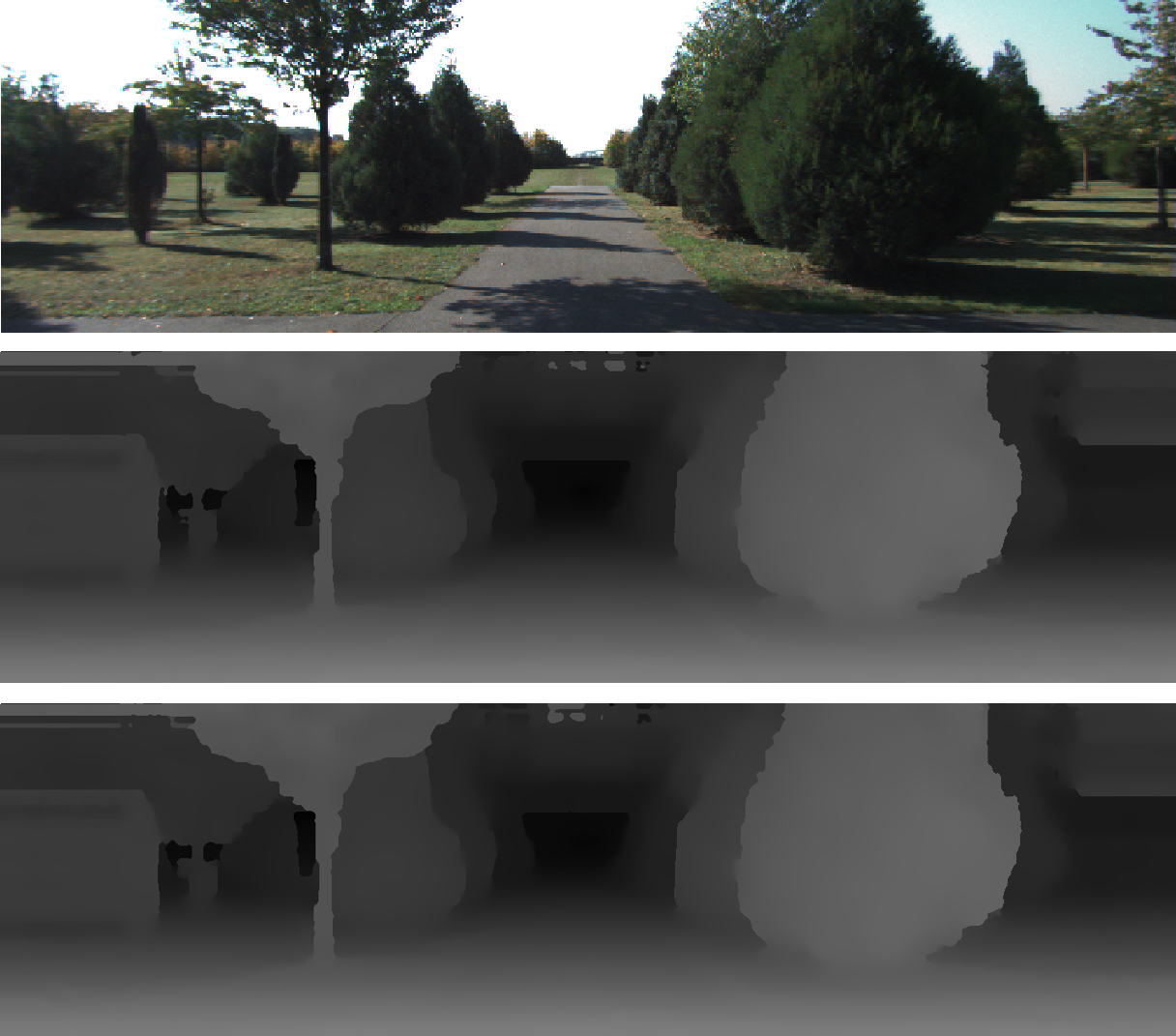}
    \caption{Visual difference between cosine similarity (center) and hamming distance (bottom) used for descriptor comparisons, the original front end from MC-CNN code base has been used. The two results are very similar, for this reason we dropped the less hardware friendly cosine similarity in favor of the more efficient hamming distance.}
    \label{fig:cosham}
\end{figure}


The first challenge we faced is the replacement of the floating point descriptor with binary ones. For this reason we took \cite{zbontar15} code-base and replaced the cosine similarity with hamming distance. Since quantization loss is to be expected we did different experiments to check how much the accuracy drops with a decreasing number of bits. What we discovered is that even using a single bit for each float number the accuracy of the output does not decrease too much. 

The way we quantize a float number into a single bit is straight forward, we simply threshold its value on zero, if the original number is bigger we set the bit to 1, 0 otherwise. We also tried to use the mean value for each output channel as a threshold but we found no improvement respect to a simple threshold on zero. Instead of 32 float feature vector (one for each output channel) we end up with an array of 32 bits. These are then compared in the Cost Matching Computation using hamming distance.
The accuracy drop we got on KITTI dataset is very small, testing for KITTI error metric T3 (all the pixel with a disparity value more wrong than 3 from the ground truth) the figures increased from 4.875\% to 4.887\%. In this way the bandwidth between the two modules decreases 118MB per frame to only 3.6MB per frame.

In Figure:~\ref{fig:cosham} it is shown the actual difference between the two results using different distance functions, the two functions give very similar results as long as the rest of the pipeline remains the same. 

\section{Ablation study}

\begin{table*}[]
\centering
\begin{tabular}{l|l|l|l|l|l|l|l|l|l|l|l|}
\cline{2-12}
                                                                                                    & Fill\_Factor(\%) & T0.125(\%) & T0.25(\%) & T0.5(\%) & T0.75(\%) & T1(\%) & T2(\%) & T4(\%) & F0.5  & F0.75 & F1.0  \\ \hline
\multicolumn{1}{|l|}{\begin{tabular}[c]{@{}l@{}}Kernel size: 3x3\\ Ch.: 1,32,32,32,32\end{tabular}} & 93.111           & 77.721     & 59.493    & 37.433   & 27.329    & 22.175 & 14.703 & 11.391 & 0.951 & 0.96  & 0.968 \\ \hline
\multicolumn{1}{|l|}{\begin{tabular}[c]{@{}l@{}}Kernel size: 3x3\\ Ch: 1,4,4,8,32\end{tabular}}     & 92.124           & 78.169     & 60.56     & 38.925   & 29.169    & 24.224 & 16.67  & 13.113 & 0.942 & 0.953 & 0.963 \\ \hline
\multicolumn{1}{|l|}{\begin{tabular}[c]{@{}l@{}}Kernel size: 9x9\\ Ch: 1,32\end{tabular}}           & 91.692           & 78.384     & 61.587    & 40.949   & 31.268    & 26.093 & 18.118 & 14.178 & 0.935 & 0.947 & 0.958 \\ \hline
\end{tabular}
\caption{In the table are shown three different network topologies, the top row represent the original architecture from \cite{zbontar15}: four layers with 32 channels each. The middle row is the smallest four layers network we tried, the bottom row is the single layer network with the input of one channel and output of 32. The dataset usd is MiddleburyV3 Quarter resolution. The metrics in the columns are respectively: the Fill Factor (which indicates the amount of non invalidated pixel), the TX metrics (which shows the percentage of disparities more wrong than X), the FX.X scores (which have the classical definition).}
\label{tab:abla}
\end{table*}

\begin{figure}
    \includegraphics[width=\columnwidth]{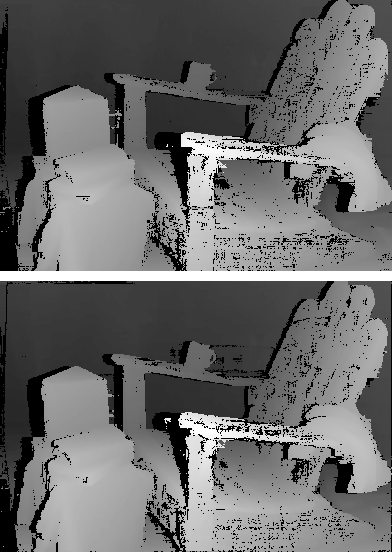}
    \caption{Visual difference between MC-CNN (top) and the single layer (bottom) network accuracies. The two results are very similar, this shows how much the fron end of the system is resilient to the neural network changes.}
    \label{fig:midd}
\end{figure}

We explored different variations of the network in Table:~\ref{tab:4l}. We first tried to decrease the number of input and output channels in different combinations of four and eight, with the only exception for the last layer which still remains 32 since we want a 32 bit descriptor. We choose four and eight mainly because the addressing and the scheduling for these values its very hardware friendly. This reduces drastically the number of FLOPs required by the network. Unfortunately the main problem with a multilayer network is the necessity to store the intermediate feature maps during the computation. Since in this use case the amount of memory is proportional to the resolution of the input image, it is very challenging to find a solution for this particular problem. To satisfy the trade off between performance and accuracy we had to make big cut on the computational budget to allow a better data flow. We decided in favor a final model with a single layer network that has 32 kernel of 9x9 size, in this way the receptive field is kept constant respect to the original model. Ablation studies were also performed in \cite{zbontar15} in susbsection \textit{Hyperparameters}, but it must be noticed that a bigger kernel size has not been tested. The usage of a single layer allows for on-the-fly computation: i.e. we don't have to wait for the whole frame to be completely processed. Essentially we can start to send binary descriptors as soon as we process 9 lines of inputs. Our experiments have shown little degradation in the final accuracy. In Figure:~\ref{fig:midd} it is shown the difference between the original MC-CNN and the single layer one. The front end (cost computation, cost aggregation, etc...) has been kept the same for both experiments. The smaller network does not degrades the accuracy dramatically and it keeps the overall metrics to an acceptable level, in Table:~\ref{tab:abla} a summary is shown.

The four metrics reported by the MiddEval3 evaluation tool are very close to the one used for the middlebury ranking. We have: "bad2.0" which measure the percentage of disparities that are at least 2 units wrong (the lower the better), "invalid" which measures the number of disparities not set by the algorithm (the lower the better), "totbad" is the total number of bad pixel (the lower the better), "avgErr" measures the distance between each predicted disparity and the ground truth (the lower the better) (Table:~\ref{tab:midtest}).

\begin{table}
\begin{tabular}{l|r|r|r|r|}
\cline{2-5}
                                       & \multicolumn{1}{l|}{bad2.0 \%} & \multicolumn{1}{l|}{invalid \%} & \multicolumn{1}{l|}{totbad \%} & \multicolumn{1}{l|}{avgErr} \\ \hline
\multicolumn{1}{|l|}{four layer}     & 16.226                           & 10.95                             & 27.178                           & 3.819                              \\ \hline
\multicolumn{1}{|l|}{single layers} & 17.39                            & 11.50                             & 28.89                            & 4.202                              \\ \hline
\end{tabular}
\caption{Final result on MiddEval3 evaluation tool.}
\label{tab:midtest}
\end{table}

In the end the number of FLOPs required to process a whole frame falls from $34\times10^9$ to a mere $265\times10^6$. This can be handled by our accelerator at 55 fps using only 10\% of it's resources.

\begin{figure*}[ht]
    \centering
    \includegraphics[width=\textwidth,height=4cm]{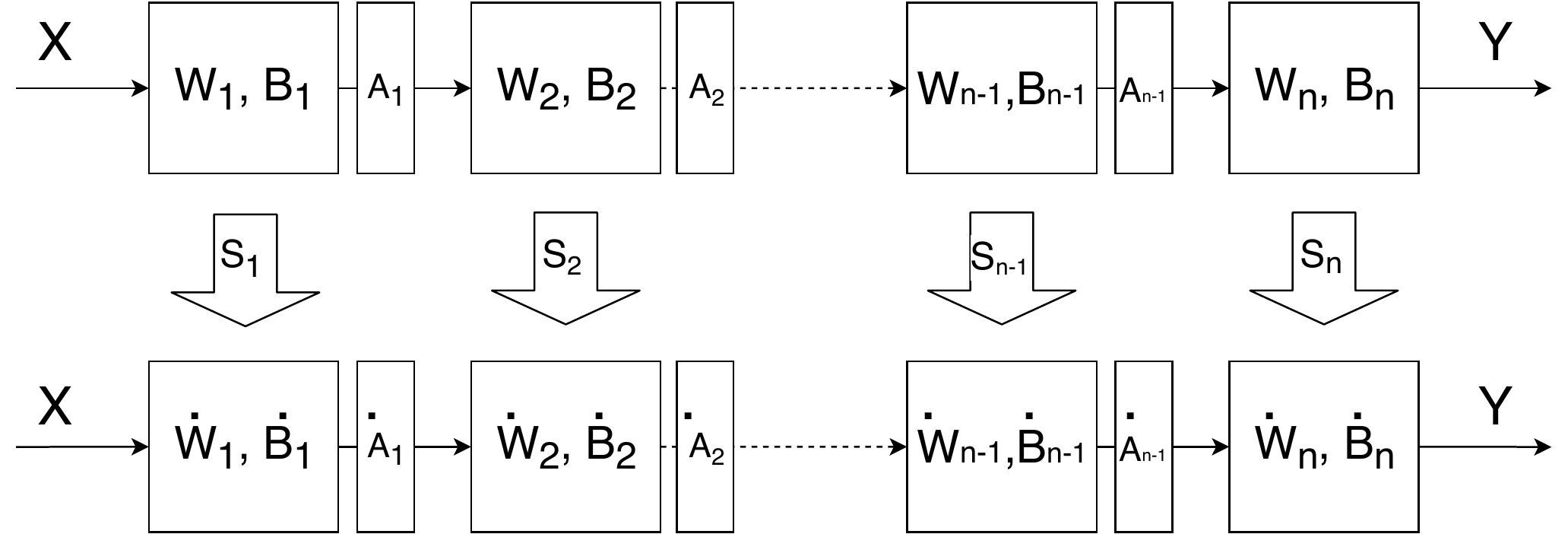}
    \caption{Quantization process: we start from a model that uses floating point for it's weights W and biases B, and searching for a suitable set of scale factors S we obtain a quantized model that has only integer weights ${\dot{W}}$ and biases ${\dot{B}}$.}
    \label{fig:quant_process}
\end{figure*}

\section{The quantization technique}

All the work we did to search for a smaller network is useless if we don't quantize its weights and features maps. For this reason we come out with the following quantization methodology, it is quite generic and works very well with shallow model like the ones we studied.
We define the original model (the one to quantize) as composed of a sequence of $L$ convolutional layer and activation layer: $model[\Gamma_1 ... \Gamma_L]$, where $\Gamma_l$ indicate the $l$-th convolution and $W_l$ and $B_l$ are respectively its weights and biases. The procedure we suggest is a serial one in which we quantize the weights and biases one layer at a time. Quantizing a convolutional layer basically means remap all the floating point weights and biases into a smaller quantized interval. The obtained set of values is indicated by $\dot{W}_l$ and $\dot{B}_l$, this new convolutional layer is defined using $\dot{\Gamma}_l$ (Fig: \ref{fig:quant_process}). The quantization interval in this case is given by the capability of our platform, i.e. 7 bits plus 1 sign bit ($int8$). A naive approach would be to spread the weights into the interval $[-128, +127]$ as much as possible, this approach doesn't give good results if the weights are not evenly distributed in the float range, e.g. if the absolute value of one of the weight is 10 times bigger than the rest. In this case the smaller weights would be quantized in the same bin. This would increase the quantization loss a lot. For this reason it is necessary to make an exhaustive search for the best division factor. Luckily, this procedure has to be performed just once in a batch manner, so the run-time is not a critical factor at this stage.

Generally speaking, the weights and biases of a convolutional layer are floating points in a narrow range (e.g. $[-1.0,1.0]$), converting them to 8 bit signed values means rescaling this interval to $[-128,127]$. We define the scale factor $S_l \in \mathbb{R}$ as the factor used to scale the weights from floating point to integer at layer $l$. The scaling is performed in the following way:

$$\dot{W}_l = round\left(\frac{W_l \cdot S_l}{\tau} \right) = R_{S_l}(W_l)$$
$$\dot{B}_l = round\left(\frac{B_l \cdot S_l}{\tau} \right) = R_{S_l}(B_l)$$

Where the $round$ function simply rounds to the nearest integer and $\tau$ is the upper limit of our interval (e.g. 127). Since $\tau$ is constant for the whole process of quantization we can redefine $S_l$ to encompass it:

\begin{equation}
\hat{S_l} = \frac{S_l}{\tau}
\end{equation}

If we now consider the input $x$, we can say that:

\begin{equation}
\begin{gathered}
\hat{S_l} \cdot \left(W_l \cdot x + B_l\right) \approx \dot{W}_l \cdot x + \dot{B}_l \\
\hat{S_l} \cdot \Gamma_l(x) \approx \dot{\Gamma}_l(x) \\
\Gamma_{\hat{S}l}(x) \approx \dot{\Gamma}_l(x)
\end{gathered}
\label{approx}
\end{equation}

Where $\Gamma_{\hat{S}l}$ is the new symbol that defines the convolution operation at layer $l$ multiplyed by $\hat{S}_l$. As you notice it is necessary to multiply the original output by the scale factor $\hat{S_l}$ to make the approximation symbol meaningful.

This approximation problem which would have been relatively easy to solve if we had only continuous function, becomes intractable in this case since the $round$ operator is introduced. Such a problem is of Integer Programming Optimization, which is NP-Complete.
This kind of complexity can be tackled only with exhaustive search algorithm or sub-optimal optimizer. We chose the first option since the interval to search is relatively small. Provided that the step is "small enough", good results can be obtained. Finally we can define our optimization problem:

\begin{equation}
\begin{gathered}
Dist(\Gamma_{\hat{S}l}, \hat{\Gamma}_l, X) \\ \\
\argmin_{S_l} \left[ Dist(\Gamma_{\hat{S}l}, \hat{\Gamma}_l, X)\right]
\end{gathered}
\label{dist}
\end{equation}

Since $S_l$ has to be positive and for $S_l$ bigger than the maximum weight there is always a big loss in quantization, we can define the search range in $(0,max(W_l)]$, with a step-search it is possible to obtain good results. In our experiments we found that if the step is three order of magnitude smaller ($<10^3$) than $max(W_l)$ we can safely assume that the search was exhaustive enough.

\subsection{Quantization loss}

There are different way to define (\ref{dist}) but the main problem is that each $S_l$ has to be tested for all possible input $X=\{x_1 ... x_n\}$. According to \cite{tensorrt} to speed up the quantization loss evaluation process it is more practical to compare the histogram of the activation instead of the actual value. Good results are claimed to be obtained for the quantization loss function called Kullback-Leibler (KL) Divergence:

\begin{equation}
D_{\mathrm{KL}}(H\|\dot{H}) = \sum_i \log_2\left(\frac{H_i}{\dot{H}_i}\right) H_i
\label{kld}
\end{equation}

Where $H$ and $\dot{H}$ are two discrete distributions which in our case are modelled upon the histograms obtained from the left and right side of (\ref{approx}). In our experiment we also tested the classic Root Mean Square Distance ($D_{RMS}$) between each respective bin:

\begin{equation}
D_{RMS}(H, \dot{H})=\frac{1}{n}\sum_i \sqrt{(H_i-\dot{H}_i)^2}
\label{rms}
\end{equation}

No relevant differences have been noticed using the two functions.

These approaches, though, rely on an important assumption: the more similar the histograms the smaller the quantization loss. This is not always true since two outputs could switch position in the histogram bin at the same time preserving the histogram shape even when the quantization function has made mistake. For this reason in our work we check the actual relation between input and output before and after the quantization. With these premises, equation (\ref{rms}) become:

\begin{equation*}
RMS\left(\Gamma_{\hat{S}l}, \dot{\Gamma}_l, X, \dot{X}\right)=\frac{1}{n}\sum_{x_i,\dot{x_i} \in X,\dot{X}}\sqrt{ (\Gamma_{\hat{S}l}(x_i) - \dot{\Gamma}_l(\dot{x}_i))^2}
\end{equation*}

Where $\dot{X}$ represents the quantized activation coming from the previously quantized layer. Using this instead of the original output allows to recover from previous error. For this reason, sometimes, during the quantization process, it is possible to see that the quantization loss decreases layer after layer. As an alternative the norm $\ell_1$ can be used:

\begin{equation*}
\ell_1 \left(\Gamma_{\hat{S}l}, \dot{\Gamma}_l, X, \dot{X}\right)=\frac{1}{n} \sum_{x_i,\dot{x_i} \in X,\dot{X}} \abs{\Gamma_{\hat{S}l}(x_i) - \dot{\Gamma}_l(\dot{x}_i)}
\end{equation*}

In our model we are going to compare binary descriptors. We check the final quantization loss using the sign of the outputs. To compute the quantization loss we used hamming distance:

\begin{equation*}
HD\left(\Gamma_{\hat{S}l}, \dot{\Gamma}_l, X, \dot{X}\right)= \sum_{x_i,\dot{x_i} \in X,\dot{X}} \sign{(\Gamma_{\hat{S}l}(x_i))} \oplus \sign{(\dot{\Gamma}_l(\dot{x}_i)})
\end{equation*}

Where the $sign$ function returns the actual sign of the output activation and then a $\oplus$ (XOR) returns 1 if the the two values have different sign, 0 otherwise.

\subsection{Channel-Wise Quantization}

If the hardware allows to set a different scale factor for each output channel of the convolution, it is possible to perform a channel-wise quantization. Allowing a different scale for each output channel can reduce the quantization loss. A different mapping for each channel onto the $int8$ range involves the post re-scaling of each component in order to make the input activation compatible with the weight of the next layer. Let $N$ be the number of output channeld of a convolutional layer $l$. Starting from (\ref{approx}) it is possible to redefine the problem for channel-wise quantization:

\begin{equation}
\begin{gathered}
\hat{S}_l = \{\hat{S}^l_1 ... \hat{S}^l_N\} \\
W_l = \{W^l_1 ... W^l_N\} \\
B_l = \{B^l_1 ... B^l_N\} \\ \\
\forall i \in [1...N] \left\{
    \begin{array}{ll}
        \hat{S}^l_i \cdot \left(W^l_i \cdot x + B^l_i\right) \approx \dot{W}^l_i \cdot x + \dot{B}^l_i \\
        \hat{S}^l_i \cdot \Gamma^l_i(x) \approx \dot{\Gamma}^l_i(x) \\
        \Gamma^l_{\hat{S}i}(x) \approx \dot{\Gamma}^l_i(x)
    \end{array} \right.\\
\end{gathered}
\label{cwq}
\end{equation}

At this point we can define Algorithm \ref{quan_conv} to quantize a single convolutional layer in a channel-wise manner.

\begin{algorithm}
  \SetKwFunction{FMain}{quant\_conv}
  \SetKwProg{Fn}{Function}{:}{}
  \Fn{\FMain{$\Gamma_l$,$X$, $\dot{X}$}}{
    $X = \Gamma_l(X)$ \;
    $\dot{X} = \Gamma_l(\dot{X})$ \;
    $\hat{S}_l = \{\hat{S}^l_1 ... \hat{S}^l_N\} $ \;
    \For{$\Gamma^l_i \in \Gamma_l$}{
        $min = \infty$ \;
        $argmin_i = 0$ \;
        \For{$s \in [0,max(W^l_i)]$}{
            $\dot{W}^l_i = round\left(W^l_i \cdot \hat{s} \right)$ \;
            $\dot{B}^l_i = round\left(B^l_i \cdot \hat{s} \right)$ \;
            \If{$Dist\left(\Gamma_{\hat{s}l}, \dot{\Gamma}_l, X, \dot{X}\right) < min$}{
                $min = Dist\left(\Gamma_{\hat{s}l}, \dot{\Gamma}_l,  X, \dot{X}\right)$ \;
                $ \hat{S}^l_i = \hat{s}$ \;
            }
        }
    }
    
    \For{$\hat{S}^l_i \in \hat{S}_l \And \dot{\Gamma}^l_i \in \dot{\Gamma}_l$}{
        $\dot{W}^l_i = round\left(W^l_i \cdot \hat{S}^l_i \right)$ \;
        $\dot{B}^l_i = round\left(B^l_i \cdot \hat{S}^l_i \right)$ \;
    }

}
 \SetKwProg{Fr}{Return}{}{}
 \label{quan_conv}
 \Fr{$\dot{\Gamma}_l$ }{} \caption{Procedure to quantize a convolutional layer}
\end{algorithm}

\subsection{Quantize Activation}

Multiplying and accumulating $int8$ values produces very large values in output (e.g. $int32$), and since the activation output of a layer has to be given as input to the next layer, it is necessary to re-scale the values to $int8$ range. Essentially, what we do after the forward pass is to reconstruct the original value of the feature map with the lowest possible quantization loss. Starting from (\ref{approx}) we get:

\begin{equation}
\begin{gathered}
    \Gamma_{\hat{S}l}(x) \approx \dot{\Gamma}_l(x) \\
    \hat{S_l} \cdot \Gamma_l(x) \approx \dot{\Gamma}_l(x) \\
    \Gamma_{l}(x) \approx \frac{\dot{\Gamma}_l(x)}{\hat{S}_l} \\
    \Gamma_{l}(x) \approx \dot{\Gamma}_{\hat{S}l}(x)
\end{gathered}
\end{equation}

Since this re-scale process is applied during the forward pass it is necessary to find an efficient way to perform the division by the $\hat{S}_l$ factor(s). Generally speaking, a multiply-shift operation is more efficient to perform compared to a division operation, especially if the multiply factor $m$ and the shift factor $h$ are integers instead of floats:

\begin{equation}
\begin{gathered}
\hat{S}_l \dot{\Gamma}_{l}(x) \approx ([\dot{\Gamma}_l(x) \cdot m] >> h) \\
\dot{\Gamma}_{\hat{S}l}(x) \approx \dot{\Gamma}^l_{m,h}(x)
\end{gathered}
\end{equation}

Where $\dot{\Gamma}^l_{m,h}$ indicate the multiply-shift operation performed after the quantized convolution at layer $l$ ($\dot{\Gamma}_l$). For this reason we optimize the following problem:

\begin{equation}
\begin{gathered}
Q(\dot{\Gamma}_{\hat{S}l},\dot{\Gamma}^l_{m,h},x) = \\ \abs{\dot{\Gamma}_{\hat{S}l}(x) - ([\dot{\Gamma}_l(x) \cdot m] >> h)} = \\
\abs{\dot{\Gamma}_{\hat{S}l}(x) - \dot{\Gamma}^l_{m,h}(x)} \\ \\
\argmin_{m,s} \{ Q(\dot{\Gamma}_{\hat{S}l}, \dot{\Gamma}^l_{m,h},x)\}
\label{mands0}
\end{gathered}
\end{equation}

Which, again, can be cast to the channel-wise quantization domain in the following way:

\begin{equation}
\begin{gathered}
Q(\Gamma^l_{\hat{S}i},\dot{\Gamma}^l_i,m_i,s_i,x) = \\ \abs{\dot{\Gamma}^l_{\hat{S}i}(x) - ([\dot{\Gamma}^l_i(x) \cdot m_i] >> h_i)} = \\
\abs{\dot{\Gamma}^l_{\hat{S}i}(x) - \dot{\Gamma}^l_{m,h,i}(x)} \\ \\
\argmin_{m_i,s_i} \{ Q(\Gamma^l_{\hat{S}i}, \dot{\Gamma}^l_{m,h,i}, x)\}
\label{mands}
\end{gathered}
\end{equation}

Where now $m$ and $h$ are two arrays containing one value for each channel $i$.

Since $m_i$ and $s_i$ are integer in a very limited range (e.g. $[0, ..., 2^z-1]$), it is not hard to check for the best pair that optimize (\ref{mands}). If we take the example of $ReLU\_127$ (a $ReLU$ with a clamping threshold of 127), the codomain of (\ref{mands}) is limited in $[-128,+127]$. It is very easy to check which is the combination that gives optimal solution (0 error). For this reason we implemented Algorithm \ref{quant_act}.

\begin{table*}[]
\begin{tabular}{l|l|l|l|l|l|l|l|l|l|l|l|}
\cline{2-12}
                                                                                                         & Fill\_Factor(\%) & T0.125(\%) & T0.25(\%) & T0.5(\%) & T0.75(\%) & T1(\%) & T2(\%) & T4(\%) & F0.5  & F0.75 & F1.0  \\ \hline
\multicolumn{1}{|l|}{\begin{tabular}[c]{@{}l@{}}Kernel size: 9x9\\ Ch: 1,32\end{tabular}}             & 91.692           & 78.384     & 61.587    & 40.949   & 31.268    & 26.093 & 18.118 & 14.178 & 0.935 & 0.947 & 0.958 \\ \hline
\multicolumn{1}{|l|}{\begin{tabular}[c]{@{}l@{}}Kernel size: 9x9\\ Ch: 1,32\\ Quantized\end{tabular}} & 91.746           & 78.387     & 61.615    & 40.993   & 31.35     & 26.158 & 18.155 & 14.21  & 0.934 & 0.947 & 0.958 \\ \hline
\end{tabular}
\caption{Accuracy drop due to the quantization loss in the single layer network. The dataset used is MiddleburyV3 Quarter resolution}
\label{tab:resq}
\end{table*}

\begin{algorithm}
  \SetKwFunction{FMain}{quant\_act}
  \SetKwProg{Fn}{Function}{:}{}
  \Fn{\FMain{$\dot{\Gamma}_{\hat{S}l},\dot{\Gamma}_l,\dot{X}$}}{
    $m = \{m_1, ..., m_N\}$ \;
    $h = \{h_1, ..., h_N\}$ \;
    \For{$\hat{S}^l_i \in \hat{S}_l$}{
        \For{$M = [0,...,2^z-1]$}{
            \For{$H = [0,...,2^z-1]$}{
                \If{$Q(\Gamma^l_{\hat{S}i}, \dot{\Gamma}^l_{M,H,i}, \dot{X}) < min$}{
                    $min = Q(\Gamma^l_{\hat{S}i}, \dot{\Gamma}^l_{M,H,i}, \dot{X})$ \;
                    $m_i = M$ \; $h_i = H$ \;
                }
            }
        }
    }
}
 \SetKwProg{Fr}{Return}{}{}
 \Fr{$m,h$ }{} \caption{Procedure to quantize the activation}
 \label{quant_act}
\end{algorithm}

Depending on the activation function (ReLU, Sigmoid, Tanh, ...) different scaling can be applied. With ReLU there are not many problems since we minimize the difference between non-quantized and quantized output, this means that positive values will stay positive and vice-versa, if the value is close to 0 the activation will be not dramatically changed. A different challenge is for Sigmoid and Tanh which have a narrow non-saturation interval, scaling this interval proportionally to $S$ solves the problem.
After that the activation and scaling is applied, the big values are again in $int8$ format.

\subsection{The algorithm}

At this point we can put all together, quantizing one layer at a time, the convolution and then the activation, it is possible to obtain very close result even if we scaled from float to $int8$.

\begin{algorithm}
  \SetKwFunction{FMain}{Quantization}
  \SetKwProg{Fn}{Function}{:}{}
  \Fn{\FMain{$model[\Gamma_1, ..., \Gamma_L],X$}}{
    $model[\dot{\Gamma}_1, ..., \dot{\Gamma}_L]$ \;
    $\dot{X} = X$ \;
    \For{$\Gamma_l \in model[\Gamma_1, ..., \Gamma_L]$}{
        $\dot{\Gamma}_l = quant\_conv(\Gamma_l, X, \dot{X})$ \;
        $\dot{\Gamma}^l_{m,h} = quant\_act(\dot{\Gamma}_{\hat{S}l},\dot{\Gamma}_l, \dot{X})$ \;
        $X = \Gamma_l(X)$ \;
        $\dot{X} = \dot{\Gamma}^l_{m,h}(\dot{X})$ \;
    }
}
 \SetKwProg{Fr}{Return}{}{}
 \Fr{$model[\dot{\Gamma}^1_{m,h}, ..., \dot{\Gamma}^L_{m,h}]$ }{} \caption{Procedure to quantize the activation}
 \label{total}
\end{algorithm}

As you can see in Algorithm \ref{total}, the input is forwarded through the original network as a reference. The final metrics reported for the choosen quantized network are shonw in Table:~\ref{tab:resq}

\begin{figure*}
    \centering
    \includegraphics[width=\textwidth]{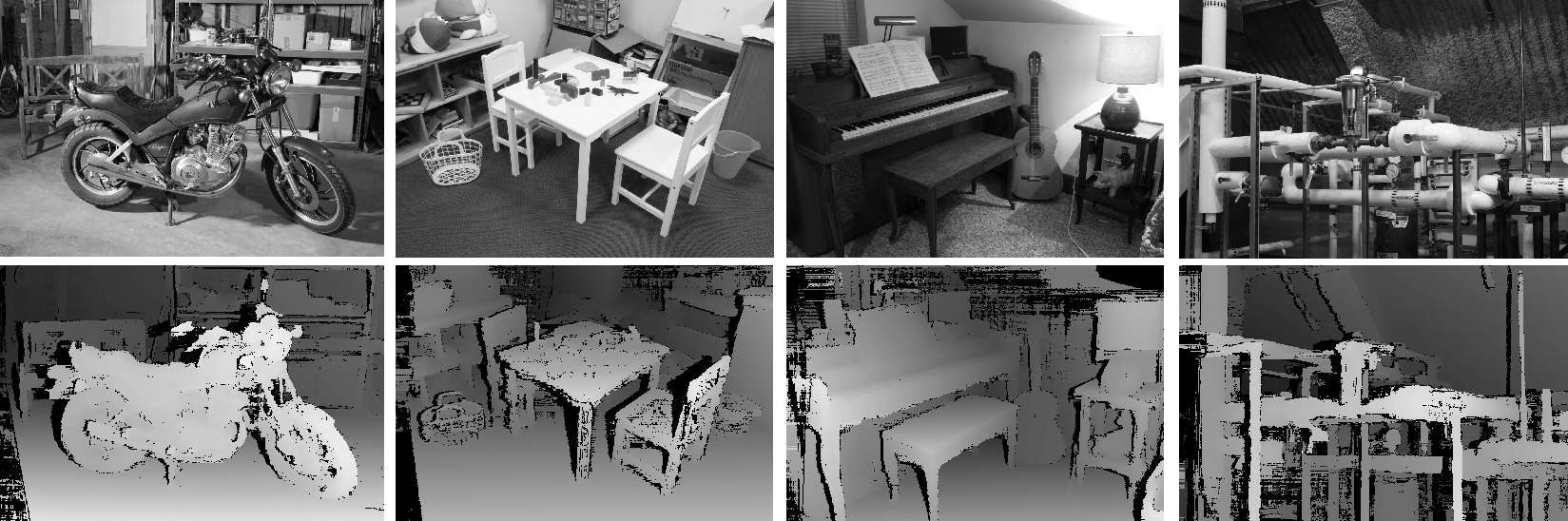}
    \vspace*{15px}
\begin{tabular}{lllllllllllll}
                                    &                             &                             &                             &                             &                             &                             &                             &                             &                             &                            &                            &                            \\ \cline{2-13} 
\multicolumn{1}{l|}{}               & \multicolumn{1}{l|}{FF}     & \multicolumn{1}{l|}{RMSv}   & \multicolumn{1}{l|}{M0.125} & \multicolumn{1}{l|}{M0.25}  & \multicolumn{1}{l|}{M0.5}   & \multicolumn{1}{l|}{M0.75}  & \multicolumn{1}{l|}{T1}     & \multicolumn{1}{l|}{T2}     & \multicolumn{1}{l|}{T4}     & \multicolumn{1}{l|}{F0.5}  & \multicolumn{1}{l|}{F0.75} & \multicolumn{1}{l|}{F1.0}  \\ \hline
\multicolumn{1}{|l|}{Middlebury} & \multicolumn{1}{r|}{92.873} & \multicolumn{1}{r|}{2.481}  & \multicolumn{1}{r|}{69.013} & \multicolumn{1}{r|}{44.932} & \multicolumn{1}{r|}{17.452} & \multicolumn{1}{r|}{6.142}  & \multicolumn{1}{r|}{23.528} & \multicolumn{1}{r|}{16.214} & \multicolumn{1}{r|}{12.663} & \multicolumn{1}{r|}{0.94}  & \multicolumn{1}{r|}{0.951} & \multicolumn{1}{r|}{0.961} \\ \hline
\multicolumn{1}{|l|}{Kitti2012}     & \multicolumn{1}{r|}{95.74}  & \multicolumn{1}{r|}{17.656} & \multicolumn{1}{r|}{78.706} & \multicolumn{1}{r|}{64.34}  & \multicolumn{1}{r|}{39.93}  & \multicolumn{1}{r|}{18.959} & \multicolumn{1}{r|}{32.35}  & \multicolumn{1}{r|}{11.886} & \multicolumn{1}{r|}{7.386}  & \multicolumn{1}{r|}{0.964} & \multicolumn{1}{r|}{0.971} & \multicolumn{1}{r|}{0.977} \\ \hline
\end{tabular}

    \includegraphics[width=\textwidth]{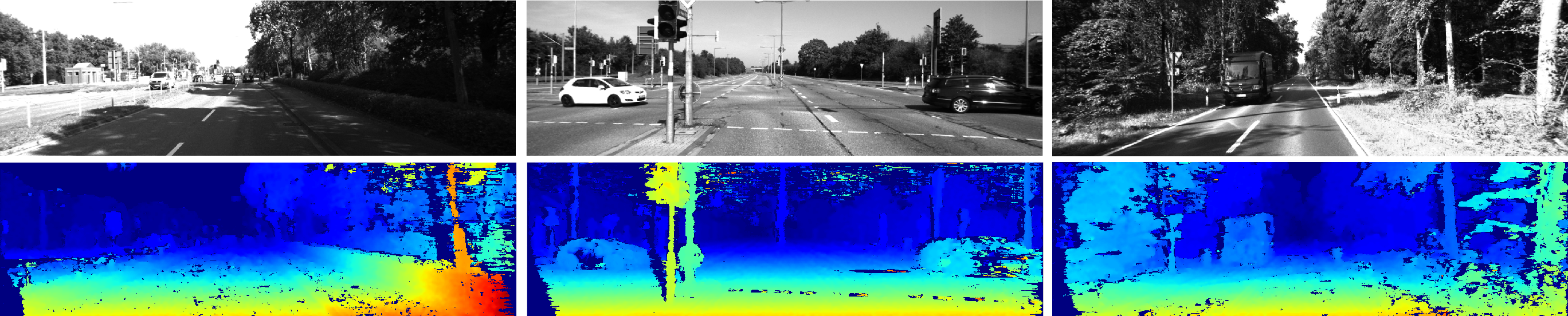}

    \caption{In figure are shown few samples taken from MiddleburyV3 Half resolution (top) and KITTI2012 dataset (bottom). The table shows different KPIs we using during the development stage: The Fill Factor indicates the amount of non invalidated pixel, RMSv is the Root Mean Square distance between the results and the ground truth performed only on the valid values, the M0.X metrics show the accuracy on the subpixel part of the result, in particulare they are only computed where the disparity is not more different than 1, the TX metrics shows the percentage of disparities more wrong than X, the F score is the classical definition.}
    \label{fig:resu}
\end{figure*}

\section{Results}

We finally compare our solution with the state-of-the-art approaches, for this reason we submitted our disparity maps to the Middlebury ranking. Currently our method ranks half way of the ranking and it is the first among all the ASICs. In Table:~\ref{tab:res} are shown the metrics we obtained:

\begin{table}
    \centering
\begin{tabular}{l|r|r|r|r|}
\cline{2-5}
                                    & \multicolumn{1}{l|}{bad0.5} & \multicolumn{1}{l|}{bad2.0} & \multicolumn{1}{l|}{rms} & \multicolumn{1}{l|}{avgerr} \\ \hline
\multicolumn{1}{|l|}{nonocc dense}  & 60.2                        & 22.8                        & 19.7                     & 6.34                        \\ \hline
\multicolumn{1}{|l|}{nonocc sparse} & 51.1                        & 15.5                        & 14                       & 3.86                        \\ \hline
\multicolumn{1}{|l|}{all dense}     & 64.4                        & 29.6                        & 29.7                     & 11.2                        \\ \hline
\multicolumn{1}{|l|}{all sparse}    & 48.6                        & 17                          & 20                       & 6.29                        \\ \hline
\end{tabular}
\caption{Results reported by the Middlebury page rank.}
    \label{tab:res}
\end{table}

The ranking measures a lot of metrics for different setup, the main difference between the setups is what it is considered for evaluation: we have a setup that considers only the non occluded pixel in the ground truth (nonocc) and one that considers all pixels. Furthermore the testing tools try to evaluate the metrics with the original results (sparse) and with a cleverly filled version (dense). Since the algorithm is capable of subpixel resolution also bad0.5 is shown. This one counts the percentage of disparities that are more wrong than 0.5 units of disparity. For a more comprehensive summary refer to Figure:~\ref{fig:resu}.

\subsection{Performance}

Since the architecture is divided in two main modules we must distinguish between the performance of the NNA and the SVA. The NNA will only run the trained CNN, this module has to process the two frames coming from the cameras. For parallelism reasons the NNA is composed of 10 submodules, using one of this submodule it is possible to process 55 frames with 720p resolution ($1280\times720$). Using only two submodules it is possible to reach more than real time performance. The SVA instead is far more efficient. It can process up to 250fps in left-right check mode. Practically speaking the module can be used for multiple camera pairs at the same time. This differentiation makes different use cases possible. While it is true that we might have the whole NNA and SVA processing stereo pairs of frame, we can also use a parte of the NNA to process a single disparity map and use the rest of the NNA to accelerate a deep learning based recognition algorithm.

\section{Conclusion}

The accuracy and performance given by our new architecture of passive Stereo Vision system is increased as never seen before in the market. Deep learning descriptors combined with a classical methodology gives an extra boost to the accuracy of the 3D reconstruction. Future development will try to make the architecture even more robust and efficient.

\bibliographystyle{plain}
\bibliography{bibliography} 





\end{document}